\pdfoutput=1 

\documentclass[11pt]{article}

\usepackage[final]{acl}

\usepackage{times}
\usepackage{latexsym}
\usepackage{url}
\usepackage{makecell}
\usepackage{booktabs}
\usepackage{multirow}
\usepackage{adjustbox}
\usepackage{tabularx}
\usepackage{listings}

\usepackage[T1]{fontenc}

\usepackage[utf8]{inputenc}

\usepackage{microtype}

\usepackage{inconsolata}

\usepackage{graphicx}
\usepackage{subcaption}
\usepackage{cleveref}

\newcommand{\daggerfootnote}[1]{%
  \begingroup
  \renewcommand\thefootnote{\fnsymbol{footnote}}%
  \footnotetext[2]{#1}%
  \endgroup
}

\title{MM-StoryAgent: Immersive Narrated Storybook Video Generation with a Multi-Agent Paradigm across Text, Image and Audio}

\author{
 \textbf{Xuenan Xu\textsuperscript{1,2}},
 \textbf{Jiahao Mei\textsuperscript{3}},
 \textbf{Chenliang Li\textsuperscript{2}},\\
 \textbf{Yuning Wu\textsuperscript{2}},
 \textbf{Ming Yan$^{\dagger}$\textsuperscript{2}},
 \textbf{Shaopeng Lai\textsuperscript{2}},
 \textbf{Ji Zhang\textsuperscript{2}},
 \textbf{Mengyue Wu$^{\dagger}$\textsuperscript{1}}
\\
 \textsuperscript{1}MoE Key Lab of Artificial Intelligence X-LANCE Lab, Shanghai Jiao Tong University\\
 \textsuperscript{2}Alibaba Group
 \textsuperscript{3}East China Normal University
\\
}

\begin{document}
\maketitle
\begin{abstract}
The rapid advancement of large language models (LLMs) and artificial intelligence-generated content (AIGC) has accelerated AI-native applications, such as AI-based storybooks that automate engaging story production for children.
However, challenges remain in improving story attractiveness, enriching storytelling expressiveness, and developing open-source evaluation benchmarks and frameworks.
Therefore, we propose and opensource MM-StoryAgent, which creates immersive narrated video storybooks with refined plots, role-consistent images, and multi-channel audio.
MM-StoryAgent designs a multi-agent framework that employs LLMs and diverse expert tools (generative models and APIs) across several modalities to produce expressive storytelling videos.
The framework enhances story attractiveness through a multi-stage writing pipeline. In addition, it improves the immersive storytelling experience by integrating sound effects with visual, music and narrative assets.
MM-StoryAgent offers a flexible, open-source platform for further development, where generative modules can be substituted.
Both objective and subjective evaluation regarding textual story quality and alignment between modalities validate the effectiveness of our proposed MM-StoryAgent system.
The demo\footnote{\url{https://huggingface.co/spaces/wsntxxn/MM-StoryAgent}}\footnote{\url{https://www.youtube.com/watch?v=2HXGrA8mg90}} and source code\footnote{\url{https://github.com/X-PLUG/MM_StoryAgent}} are available.
\end{abstract}

\section{Introduction}
\daggerfootnote{Corresponding author.}

\begin{figure}[t]
   \centering
   \includegraphics[width=\linewidth]{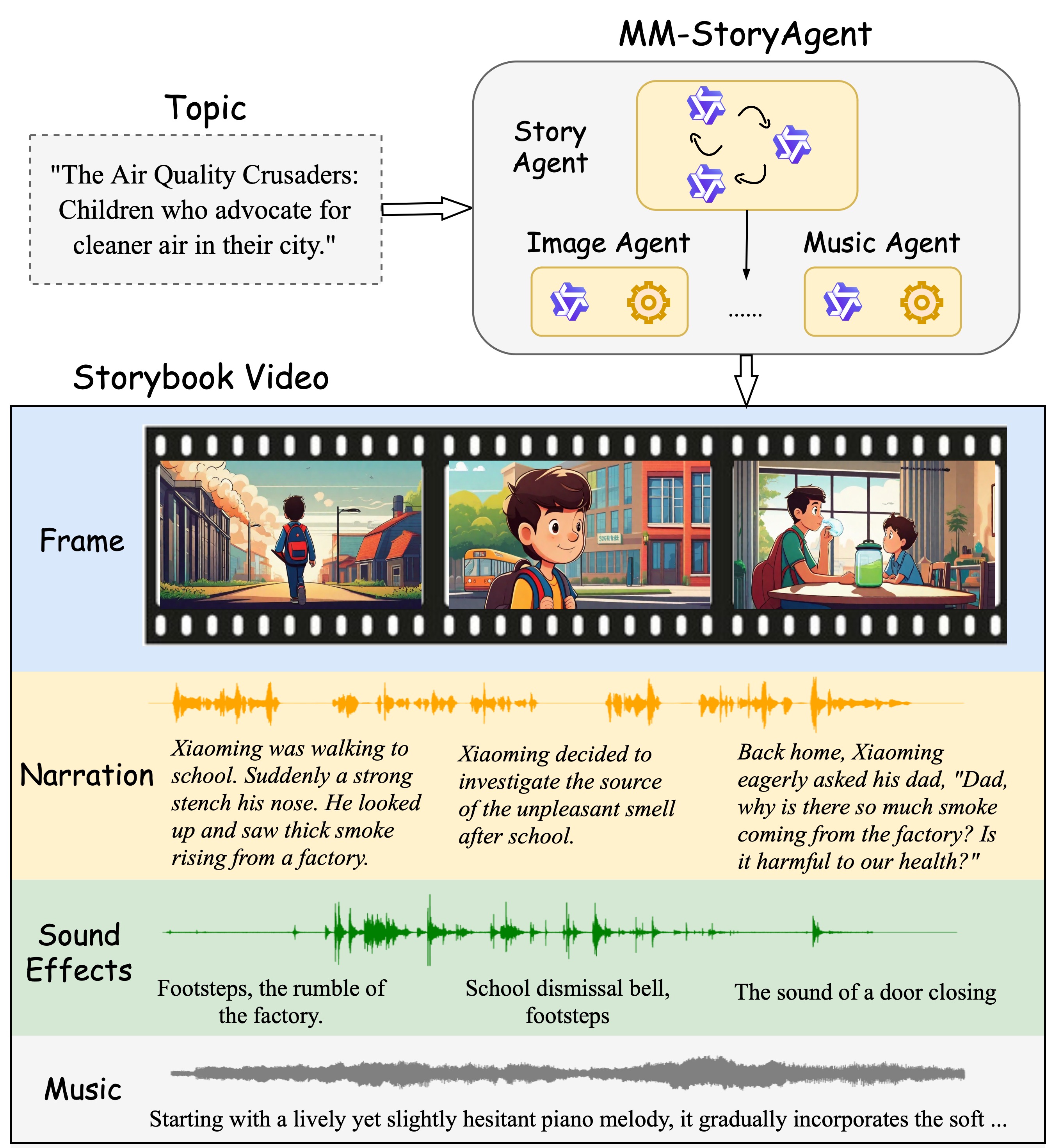}
   \caption{MM-StoryAgent incorporates LLMs, generative models and tools to transform the input story setting into omni-modality storytelling videos.}
   \label{fig:simple_diagram}
\end{figure}

The success of generative models enables the automatic creation of high-quality media content from several modalities.
In the textual modality, large language models (LLMs) demonstrate remarkable proficiency in generating coherent and contextually relevant pieces of text while following human instructions.
In the visual modality, text-to-image~\cite{ldm} and text-to-video~\cite{imagen_video} models achieved impressive results in generating high-quality images and videos from textual descriptions.
In the audio modality, high-fidelity samples of various audio types, e.g., speech~\cite{valle}, sound~\cite{audioldm2}, and music~\cite{musicgen}, can be generated.
Such advancements of AIGC has given rise to plenty of products and applications, among which AI-based storybooks~\cite{StoryCom,StorybookAI} are attracting increasing attention.
The utilization of AIGC tools for storybook generation saves the time consumed in generating diverse stories in traditional production processes, providing accompaniment experiences for children. 

Despite these advancements, the automatic generation of narrated storybooks still faces significant challenges.
\textbf{First, the generated story quality needs improvement.}
The plot attractiveness and theme depth are core traits of a good story.
However, LLM-generated stories are not attractive enough.
These stories are usually flat and straightforward, lacking in tension and conflict, which makes them less engaging for readers.
\textbf{Second, the presentation modalities need to be further enriched.}
Most efforts on story generation focused on visual or textual modalities.
However, there is a growing expectation for a more holistic multimodal expression, across image, text and multi-channel audio, to enhance the overall immersive quality of the storytelling experience.
\textbf{Third, an open-source, flexible storytelling video generation framework and evaluation benchmark are lacking.}
Currently, most AI storybook generation works are closed-source applications.
Moreover, there is a lack of publicly available datasets and evaluation standards for assessing story quality.


\begin{figure*}[ht]
   \centering
    \includegraphics[width=0.95\linewidth]{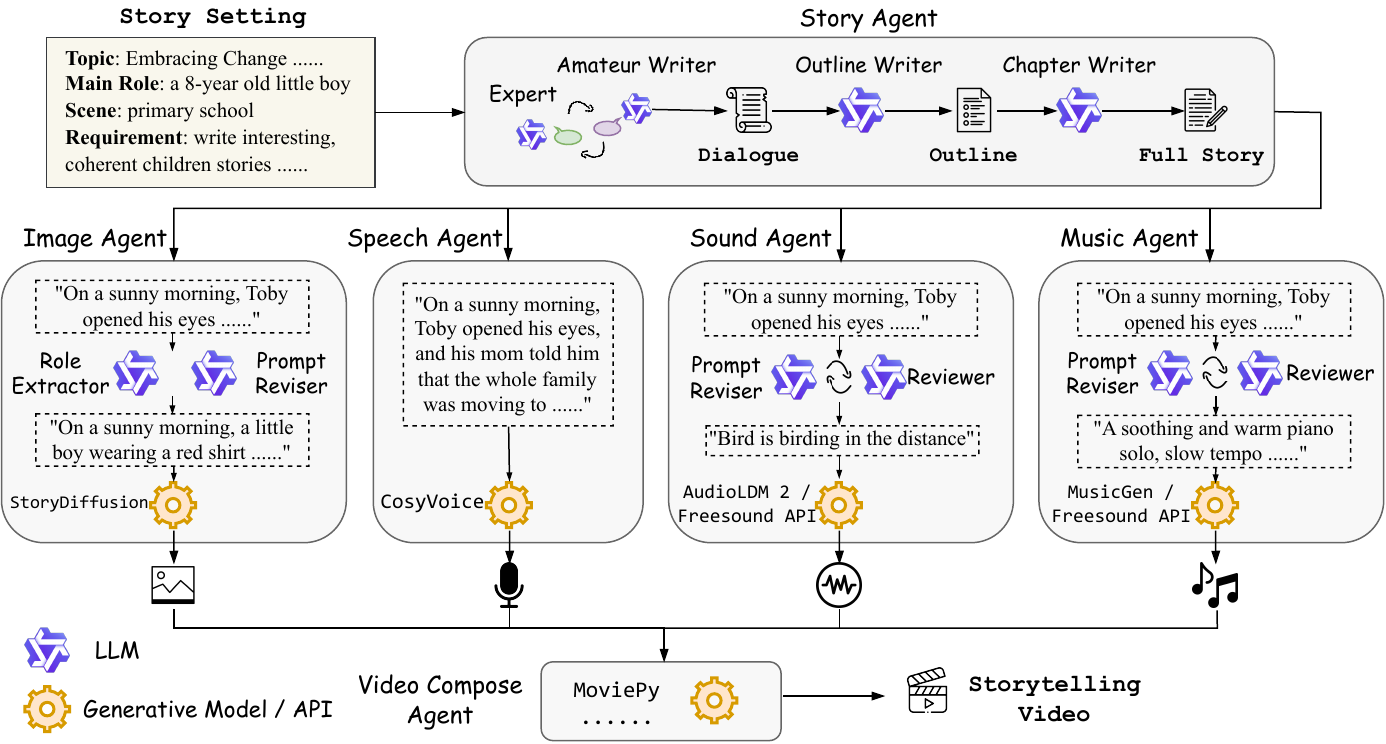}
    \caption{The overview of MM-StoryAgent framework. Agents discuss the writing requirements given story settings before the outline-chapter two-stage writing process. Then modality-specific assets are produced by generation tools and potential prompt revisers. Finally, the video compose agent produces the storytelling video.}
    \label{fig:framework}
\end{figure*}

To address these challenges, we propose \textit{MM-StoryAgent}, a \textbf{M}ultimodal, \textbf{M}ulti-agent framework for narrated storybook video generation.
MM-StoryAgent leverages the capabilities of LLMs and expert tools of several modalities to make the narrated storybook video, as illustrated in \Cref{fig:simple_diagram}.
First, in the story generation process, we adopt a multi-agent, multi-stage story writing pipeline to improve the attractiveness of story.
The dialogue between agents provide ideas on writing attractive stories.
Evaluation on a series of topics validates the effectiveness of the writing pipeline.
Then, we integrate image and multi-channel audio (speech, music and sound effects) to provide omni-modality experience.
To align the textual story content with different modalities, multiple agents are designed to revise the original story to modality-specific prompts.
Reflection is adopted to enhance the quality of the revised prompt.
In terms of image, role extraction agents and image prompt revisers work separately to achieve the role consistency.
In terms of audio, speech, music and sound effects are generated separately and mixed to produce multi-channel audio.
The composition of omni-modality assets improve the presentation richness of storybooks.
Finally, we propose a flexible and modularized framework agents for narrated storybook video generation and an evaluation topic dataset with corresponding criteria.

Overall, we make the following contributions to promote advances in narrated storybook video generation:
\begin{itemize}
    \item \textbf{Improved Story Quality: }MM-StoryAgent enhances the attractiveness of generated stories by adopting a multi-agent, multi-stage writing pipeline. 
    \item \textbf{Omni-Modality Presentation of Story: }MM-StoryAgent integrate images and multi-channel audio (narration speech, music and sound effects) to provide an immersive storytelling experience across full modalities. 
    \item \textbf{Opensourcing Framework and Evaluation: }MM-StoryAgent provides a flexible and modularized story video generation framework, as well as a topic dataset with criteria for story generation evaluation.  
\end{itemize}

\section{MM-StoryAgent}

\subsection{Framework Overview}

MM-StoryAgent is a multi-agent storytelling video generation framework.
As shown in \Cref{fig:framework}, the workflow incorporates multiple agents, including LLMs, generative models and APIs, as well as the communication and collaboration between them.
The whole framework takes the story setting as the input.
The story agent adopts a multi-agent, multi-stage writing pipeline to write high-quality stories.
The dialogue on writing ideas before the formal writing stage enhances the plot richness, making the story more engaging. 
Then, due to the gap between the story and the modality focus, modality-specific agents transform the raw story content into prompts by multi-agent communication.
A prompt reviser is collaborating with a reviewer to iteratively polish the prompt. 
Generative models or search APIs are called to generate corresponding assets.
Finally, assets of all modalities are composed to produce the full storytelling video with immersive reading experience.

Detailed workflow of each agent is presented in the following part.
These workflows are independent of each other so assets of all modalities can be generated in parallel to accelerate the final video production.
Data are passed between agents in a structured JSON format to standardize the interaction interface between agents.
In addition, the implementation of each agent is flexible and not bound with specific models.
For example, any text-to-speech (TTS) model is compatible with the speech agent.

\subsection{Attractiveness-Oriented Story Agent}
\paragraph*{Question-Answering Dialogue Simulation}
Given the story setting, we simulate a dialogue between an amateur story writer and an experienced story writing expert.
They discuss various aspects of considerations for story writing to meet the writing requirements, such as appropriate role characteristics.
Such a dialogue simulates the brainstorming process that humans go through before they start writing, allowing the agent to clarify how to make stories more attractive.
Hereby, writing guidance is provided for the agent so that the difficulty of the original writing task is reduced.
In each question-answering turn, we use the story setting and the dialogue history to prompt the writer to ask questions so that the writer obtains deeper understanding of the writing requirements.
The dialogue continues for at most $T_d$ turns.

\paragraph*{Outline Writing}
With the deep discussion on the writing requirements, the simulated dialogue often contains detailed information of the story, e.g., the general plot arc.
Therefore, we prompt an outline writing agent to write the outline based on the dialogue.
Different from STORM~\cite{storm}, we do not use the dialogue to refine an outline draft.
Since the dialogue provides detailed information that often deviates significantly from the draft outline, it is difficult for the outline writer to refine the draft with full consideration of the information in the dialogue.

\paragraph*{Chapter Writing}
After generating the story outline, the story agent calls a chapter writer to expand the outline to the full-length story chapter-by-chapter.
The expansion supplements rich details to the brief outline, making the development and transitions of the story plots more reasonable and natural.
To avoid the incoherence caused by parallel outline-to-chapter expansion, chapters are generated sequentially.
The previously generated chapters are given when generating the current chapter.
Finally, all generated chapters are concatenated to get a coherent story adhering to the use-provided story setting.

\subsection{Modality-Specific Agents}
Based on the generated attractive story, modality-specific agents generate corresponding assets to provide omni-modality storybook reading experience.

\subsubsection{Image Agent}
\paragraph{Visual Description Extraction}
The image agent is responsible for generating illustrations for each chapter of the story.
The original story content focuses on the plot, while the text-to-image generation model prefers descriptions focusing on visual elements.
To fill this gap, we employ an agent to transform the story content into image prompts, where non-visual information such as sound and psychological activity is eliminated.
In addition, the role consistency is an important requirement for story illustrations.
We improve the consistency by using consistent role descriptions in the image prompts.
This is accomplished by another role extractor agent.
It summarizes all main roles in the story and generate concise descriptions for each role.
Then, the role name in transformed image prompts are replaced by corresponding descriptions to be fed to the generation model.

Both the role extraction and image prompt revision processes follow a ``reviser-reviewer'' multi-agent multi-turn diagram.
Given the instruction, in each turn the writer generates the outcome and the reviewer checks whether the outcome meets requirements in the instruction.
If requirements are not met, the reviewer provides specific reviews for the writer to refine the outcome in the next turn.
Such a process continues until the reviewer decides that requirements are met or it reaches a maximum turn number $T_p$.
The reflection from reviewer helps the reviser adhere to the revising instruction.

\paragraph{Image Generation} With the visual-centered prompts, we utilize StoryDiffusion~\cite{storydiffusion} as the generative model.
StoryDiffusion replaces the original self-attention with a new consistent self-attention so that the model generates the next image frame with the attention to previous frames to enhance the role consistency.
We use Stable Diffusion XL as the generation backbone.

\subsubsection{Audio Agents}
\label{subsubsec:audio_agents}
Three channels of audible content are generated: speech, sound and music, which will be processed separately.
Here, sound refers to non-speech, non-music sound effects.
\paragraph*{Speech Agent}
The workflow of the speech agent is straightforward: it directly transforms the story content into the narrated speech.
We use open-source CosyVoice~\cite{cosyvoice} as the TTS agent.

\begin{table*}[ht]
    \centering
    \small
    \begin{tabular}{c|c|c|c|c|c|c|c|c|c}
    \toprule
    \multicolumn{2}{c|}{\multirow{2}{*}{Objective Score}} & \multicolumn{3}{c|}{Textual Story} & \multicolumn{5}{c}{Modality Alignment} \\
    \cline{3-10}
    \multicolumn{1}{c}{} & & A & W & E & I-T & S-T & M-T & I-S & I-M \\
    \midrule
    \multirow{2}{*}{Topic 1: Self-growing} & Direct & 3.80 & 4.28 & 3.72 & 0.297 & 0.224 & 0.324 & 0.0574 & 0.0434 \\
     & Story Agent & 3.72 & 4.32 & 3.75 & 0.324 & 0.232 &  0.542 & 0.0548 & 0.0540 \\
    \midrule
    \multirow{2}{*}{Topic 2: Family \& Friendship} & Direct & 3.80 & 4.60 & 3.28 & 0.298 & 0.192 & 0.266 & 0.0482 & 0.0359 \\
     & Story Agent & 4.04 & 4.68 & 3.68 & 0.313 & 0.253 & 0.491 & 0.0441 & 0.0361 \\
    \midrule
    \multirow{2}{*}{Topic 3: Environments} & Direct & 3.92 & 4.32 & 3.76 & 0.293 & 0.201 & 0.288 & 0.0542 & 0.0410 \\
     & Story Agent & 4.04 & 4.32 &  3.88 & 0.309 & 0.216 & 0.529 & 0.0492 & 0.0536 \\
    \midrule
    \multirow{2}{*}{Topic 4: Knowledge Learning} & Direct & 3.68 & 3.52 & 3.56 & 0.300 & 0.238 & 0.344 & 0.0562 & 0.0482 \\
     & Story Agent & 3.96 & 3.52 & 3.84 & 0.320 &  0.260 & 0.538 & 0.0493 & 0.0534 \\
    \midrule
    \multirow{2}{*}{All} & Direct & 3.80 & 4.18 & 3.58  & 0.297 & 0.214 & 0.301 & \textbf{0.054} & 0.0421 \\
     & Story Agent & \textbf{3.94} & \textbf{4.21} & \textbf{3.79} & \textbf{0.316} &  \textbf{0.240} & \textbf{0.525} & 0.0494 & \textbf{0.0493} \\
    \bottomrule
    \end{tabular}
    \caption{Objective evaluation of the textual quality and modality alignment of generated storytelling videos. The textual quality in terms of attractiveness (A), warmth (W) and education (E) are graded by a 1-5 scale of rubric grading. Modality alignment scores measure the content alignment between image (I), sound (S), music (M) and text (T).}
    \label{tab:objective_evaluation}
\end{table*}

\paragraph*{Sound / Music Description Extraction}
Workflows of the sound and music agent are similar to the image agent, both consisting of a ``reviser-reviewer'' iterative prompt revision stage and a generation stage.
Prompts focusing on sound effects are extracted from the original story content to eliminate non-audible elements.
Foreground speech and all music descriptions are also excluded since distinguishable speech and background music are generated by other agents.
For music, the prompt reviser analyzes the story setting and content to summarize the appropriate background music description, including instruments, styles, etc.

\paragraph{Sound / Music Generation}
We use AudioLDM 2~\cite{audioldm2} to generate sound effects.
Since the performance of current sound generation models may be less satisfactory when handling prompts that are rarely seen during training~\cite{chang2024open}, the retrieval approach is also provided as a candidate.
Each sound effect type can be retrieved via Freesound API~\cite{font2013freesound}.
MusicGen~\cite{musicgen} is taken to generative music.
To avoid unstable generation result, we can also use music retrieved by APIs (e.g., Freesound search API).

\subsection{Video Composition Agent}
After assets of all modalities are generated, they are composed into an omni-modality video: temporal images corresponding to different story pages, audible content including narration of the story, sound effects and background music.
The composition of assets is based on the duration of the narrated speech.
Each image frame is displayed for the corresponding speech with a duration of $t$ seconds, while the associated sound is also repeated or truncated to $t$ seconds accordingly.
Finally, the background music is played all the time.
To achieve a more natural display and transition effect, each page incorporates random slow movement or zoom in / out effects.
Furthermore, slide in / out transitions are applied between pages.
The composition functionality is implemented by libraries such as MoviePy.

\section{Experiments and Evaluation}
To validate the effectiveness of MM-StoryAgent, we perform both automatic objective evaluation and subjective human evaluation.
The quality of the textual story and the alignment between the modalities are assessed.
The baseline method is a direct prompting approach (\textit{Direct}), where the LLM is directly prompted to write a story given the setting, and the generated textual story is also directly fed to AIGC tools to generate modality assets.
In comparison, \textit{Story Agent} simulates a dialogue to discuss how to follow the setting before the two-stage outline-story writing.
We use \texttt{qwen-2-72b-instruct} as the LLM and $T_d$ is set to 3.

\subsection{Objective Evaluation}

\paragraph*{Textual Story}

We select 4 common children storybook topic types: \textit{self-growing}, \textit{family and friendship}, \textit{environments} and \textit{knowledge learning}.
For each topic type, we prompt the LLM to list 25 story topics, resulting in a total of 100 topics.
We use \texttt{gpt-4} to score the story based on a 5-point rubric grading on 3 aspects: \textit{attractiveness}, \textit{warmth} (the warmth and emotion conveyed by the story), and \textit{education} (the educational content conveyed by the story).
These criteria are the most distinguished features of good children stories.


Results are presented in the left half of \Cref{tab:objective_evaluation}.
The overall results show that Story Agent exhibits substantial improvement.
Story Agent outperforms the Direct baseline in adhering to the provided instructions.
For example, the topic ``family \& friendship'' is less associated with education so Direct gets a lower education score, indicating the lack of educational content. 
In contrast, Story Agent pipeline can follow the writing instruction to convey educational content better.

\paragraph*{Prompt Revision}

For prompt revision, $T_p$ is set to 3 in the revision stage.
We first use objective cross-modal alignment scores from contrastive learning models to assess the effect of prompt revision.
We conducted modality alignment evaluations across 100 story topics using the following metrics:
1) CLIP~\cite{clip} score evaluates the image-text (I-T) alignment between the generated image and its corresponding text prompt.
2) CLAP~\cite{clap} score measures the alignment between the generated sound/music (S-T/M-T) and its corresponding text prompt.
3) Wav2CLIP~\cite{wav2clip} score measures the alignment between image and sound/music (I-S/I-M) by calculating audio-visual modality similarity scores.
Results demonstrate that the Story Agent significantly outperforms the Direct baseline across all dimensions except for image-sound alignment.
This highlights the effectiveness of prompt revision in improving the alignment between the generated assets and the story content, thereby enhancing the overall quality of the generated videos.
As stated before, sound generation models tend to struggle when handling complex prompts while LLMs often generate prompts with rich descriptions, even when explicitly instructed to keep prompts concise.
This mismatch may result in suboptimal sound effects produced by Story Agent.
The improvement in sound generation models and the instruction following capability of LLMs may mitigate this problem in the future.


\subsection{Subjective Evaluation}

\begin{table}[ht]
    \centering
    \small
    \begin{tabular}{c|c|cc}
    \toprule
    \multicolumn{2}{c|}{Method} & Direct & Story Agent \\
    \midrule
    \multirow{3}{*}{\makecell{Textual\\Story}} & Attractiveness & \textbf{3.87} & \textbf{3.87} \\
    & Education & 3.63 & \textbf{3.80} \\
    & Warmth & 3.70 & \textbf{3.87} \\
    \midrule
    \multirow{3}{*}{\makecell{Modality\\Alignment}} & Sound & \textbf{2.93} & 2.53 \\
    & Music & 2.57 & \textbf{2.93} \\
    & Image & 2.77 & \textbf{3.47} \\
    \bottomrule
    \end{tabular}
    \caption{Subjective evaluation of textual quality and modality alignment of storytelling videos.}
    \label{tab:subjective_evaluation}
\end{table}

We invite three well-educated raters to conduct subjective evaluations on text story quality and modality alignment.
Raters are all familiar with media content and children storybooks.
They are asked to rate each aspect of videos on a 1-5 scale.
20 videos across the four topics generated by Story Agent and Direct methods are evaluated.
An agreement score of 0.46 is reached, indicating inter-rater moderate agreement in spite of the limited subjective evaluation scope.

\Cref{tab:subjective_evaluation} demonstrates evaluation results.
Across all dimensions of textual story quality, Story Agent consistently outperforms the Direct baseline.
This demonstrates that the multi-agent, multi-stage story writing pipeline significantly improves story quality, aligning with objective evaluation results.
For modality alignment, the subjective evaluation also aligns with the objective scores.
Specifically, improvements in music and image alignment highlight the effectiveness of Story Agent in employing prompt revisers and reviewers to refine prompts, enhancing the overall match between modalities and the story content.
However, for sound, the prompt reviser appears to introduce excessive or less suitable sound effects, resulting in unwanted sound elements.
This also represents an area for future improvement.

\section{Related Work}
MM-StoryAgent is based upon the latest advancement of LLMs and generative models of different modalities.
This section gives a brief overview of them.

\paragraph{Large Language Models}
Large language models exhibit remarkable generalization capabilities across various challenging language tasks~\cite{llama,qwen2}, such as coding and role playing.
Such advancements enable automatic full-length article writing following specific literary styles that previously performed poorly with small models.
Recently, a rising efforts have been directed towards story-like narrative creation tasks~\cite{Mirowski2022CoWritingSA, ma2024mopsmodularstorypremise,storm}. For example, STORM~\cite{storm} designed a Wikipedia-like article writing system. Weaver~\cite{wang2024weaver} pretrains and finetunes a foundation LLM for creative writing. 


In addition to accomplish writing tasks, LLMs have also been integrated to develop agents to utilize various tools to accomplish complex tasks~\cite{shen2024hugginggpt,hong2023metagpt,li2023modelscope}.
For example, HuggingGPT~\cite{shen2024hugginggpt} and Autogen~\cite{wu2023autogen} take LLMs as the brain for planning and decomposing complex user requests into simple tasks that can be solved by calling tools.
Inspired by these applications, we utilize LLMs to serve as a bridge the story content and generative models of all modalities within the storybook video generation framework.

\paragraph{Visual and Audio Generative Models}
As auto-regressive models demonstrated success in discrete text generation, generative models~\cite{vae,gan,nflow} for continuous data have also seen rapid developments.
Among these generative models, diffusion models~\cite{ddpm} are distinguished for their abilities to generate diverse and high-quality data samples.
Therefore, diffusion models have become the mainstream model in image~\cite{ldm}, video~\cite{imagen_video} and audio~\cite{gradtts,audioldm,noise2music} generative modelling.
For the application of story-type image generation, some efforts have been made to enhance the consistency of characters across multiple frames~\cite{storydiffusion,he2024dreamstory}.
In MM-StoryAgent, we choose these models as modality experts to generate single-modality elements, which are further composed to make long-duration narrated story videos.
With better generative models proposed, these experts can be replaced by more advanced models to continuously improve the video quality.

\section{Conclusion}

In this work, we propose MM-StoryAgent, a multi-agent framework aimed at advancing narrated storybook video generation.
By employing a multi-agent, multi-stage writing pipeline, MM-StoryAgent significantly enhances the quality of the generated stories.
Moreover, the integration of visual, auditory, and narrative elements creates a more immersive storytelling experience.
MM-StoryAgent is also the first open-source framework for narrated storybook video generation that supports flexible model and API integration.
To validate the effectiveness of MM-StoryAgent, we propose a story topic list and corresponding story evaluation criteria.
Automatic evaluation on story quality and modality alignment show the advantage of our MM-StoryAgent framework.

\newpage

\section*{Limitations}

MM-StoryAgent focuses on improving story quality and bridging the gaps between various modalities and story content.
However, the alignment between modalities requires further exploration.
For example, the specific timing of short-duration sound effects should be aligned with the narration.
Additionally, the effects of tools, such as text-to-image generation models, are key limiting factors of the final video quality.

\section*{Ethics Statement}

Although MM-StoryAgent is intended to generate diverse and attractive stories, it is not perfect.
It is still possible that MM-StoryAgent generate biased or illogical stories due to the inherent bias of agents.
Users may add a detection process in the real application.

\section*{Acknowledgments}
This research was funded by Alibaba Innovative Research Project and Guangxi Major Science and Technology Project (No. AA23062062).

\bibliography{custom}

\clearpage

\appendix

\section{Coherence and Relevance Evaluation}

\begin{figure*}[ht]
    \centering
    \includegraphics[width=\linewidth]{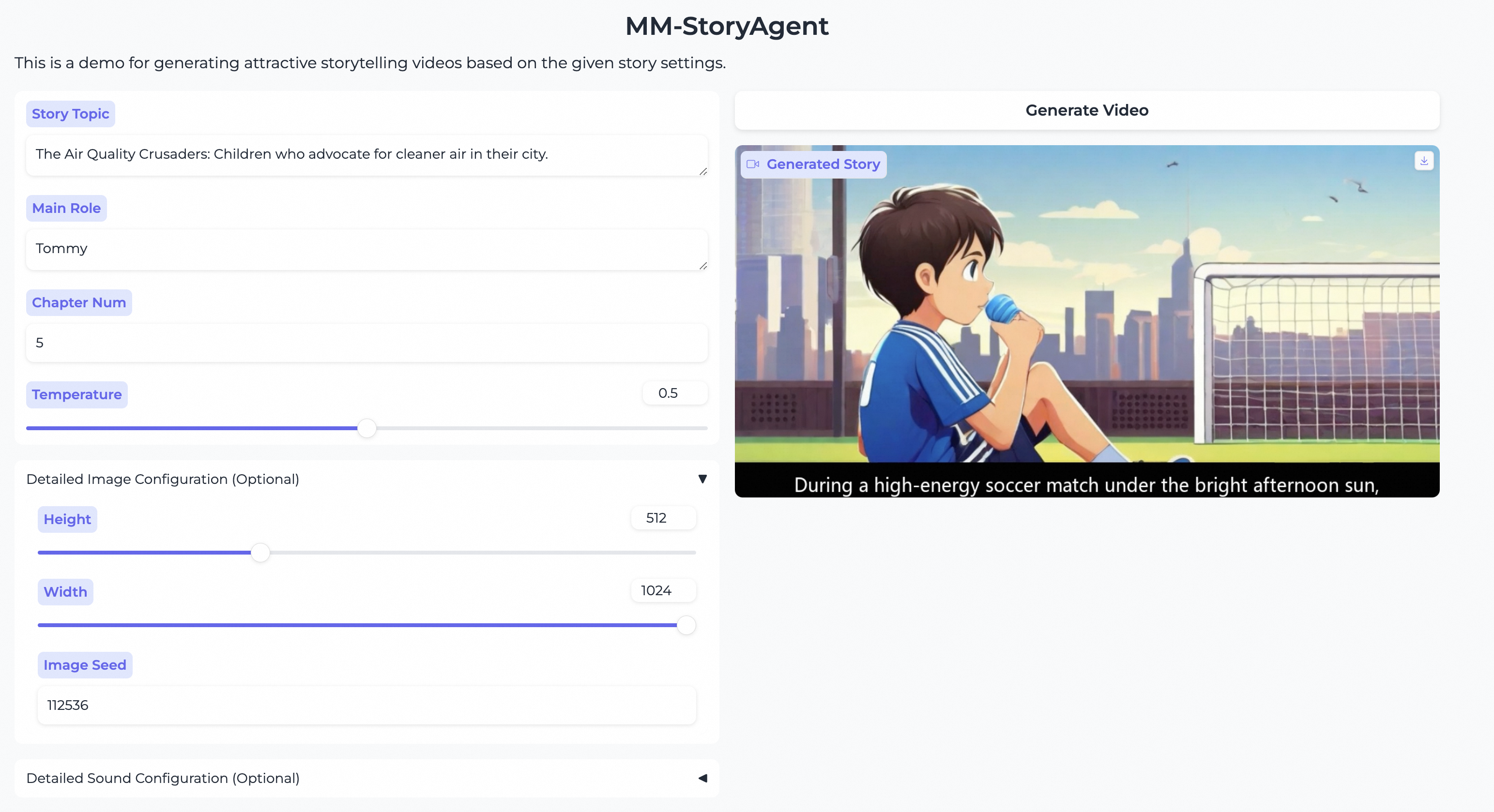}
    \caption{The gradio demonstration page.}
    \label{fig:gradio_demo}
\end{figure*}

\label{sec:coherence_relevance_eval}

In addition to the most distinguished story features (attractiveness, warmth, education), we evaluate the story quality in terms of two basic writing requirements: 1) \textit{relevance}: is the story relevant to the given topic? 2) \textit{coherence}: is the story logically coherent?
These aspects are included as requirements in the story settings of both Direct and Story Agent approaches.
Both the Direct and Story Agent approaches demonstrate promising performance, with coherence scores around 4.4 and relevance scores approaching 5 (see \Cref{tab:story_quality_eval_appendix}).
The subtle decline in coherence using Story Agent may be attributed to the fact that Story Agent adopts an outline-story two-stage writing pipeline, which places higher demands on the instruction-following capability of the foundation model.
Evaluation results on all aspects indicate that stories created by the Story Agent can enhance the distinguished story features while ensuring the basic story requirements.

\begin{table}[ht]
    \centering \small
    \begin{tabular}{c|c|c|c}
    \toprule
    \multicolumn{2}{c|}{Rubric Grading} & Relevance & Coherence \\ 
    \midrule
    \multirow{2}{*}{Topic 1} & Direct & 4.60 & 4.44 \\
     & Story Agent & 4.76 & 4.27 \\
    \midrule
    \multirow{2}{*}{Topic 2} & Direct & 4.88 & 4.48 \\
     & Story Agent & 4.88 & 4.44 \\
    \midrule
    \multirow{2}{*}{Topic 3} & Direct & 5.00 & 4.44 \\
     & Story Agent & 4.96 & 4.40 \\ 
    \midrule
    \multirow{2}{*}{Topic 4} & Direct & 4.92 & 4.24 \\
     & Story Agent & 4.96 & 4.28 \\
    \midrule
    \multirow{2}{*}{All} & Direct & 4.85 & \textbf{4.40} \\
     & Story Agent & \textbf{4.89} & 4.35 \\
    \bottomrule
    \end{tabular}
    \caption{Results of automatic story quality evaluation. The rubric grading uses a 1-5 scale.}
    \label{tab:story_quality_eval_appendix}
\end{table}

\begin{table}[t]
    \renewcommand{\arraystretch}{1.2}
    \small
    \begin{adjustbox}{width=\columnwidth}
        \begin{tabularx}{\columnwidth}{X}
        \hline
            \hline
            \textbf{Prompt for Story to Image Description Conversion} \\
            \hline
            Convert the provided story content into image descriptions. If there are previous results and improvement suggestions, modify the image description based on the feedback. \newline
            \textbf{Input Format:} \\
            Input consists of all story content, the current story content, and possibly the previous output and corresponding feedback, formatted as: \\
            \{\\
            \hspace{2em} "all\_stories": ["xxx", "xxx"], // Each element represents a page of story content. \\
            \hspace{2em}    "current\_story": "xxx", \\
            \hspace{2em}    "previous\_result": "xxx", // Empty means the first round.\\
            \hspace{2em}    "improvement\_suggestion": "xxx" // Empty means the first round.\\
            \}\\
            \textbf{Output Format:} \\
            Output a string representing the image description corresponding to the current story content. \\
            \textbf{Notes:} \\
            1. Keep it concise. Describe only the main elements, omitting details. \newline
            2. Retain visual elements. Only describe static visuals, avoiding any narrative. \newline
            3. Remove non-visual elements. Typical non-visual elements include dialogue, psychological activities, and plot. \newline
            4. Retain character names. \\
            \hline
        \end{tabularx}
    \end{adjustbox}    
    \caption{\label{tab:prompts-story-to-image} Prompts for story to image description conversion.}
\end{table}

\begin{table}[t]
    \renewcommand{\arraystretch}{1.2}
    \small
    \begin{adjustbox}{width=\columnwidth}
        \begin{tabularx}{\columnwidth}{X}
        \hline
            \hline
            \textbf{Prompt for Image Description Review} \\
            \hline
            Review the provided image description corresponding to the story plot. \\
            If it meets the requirements, output "Check Passed".\\ If not, provide improvement suggestions. \newline
            \textbf{Requirements for Image Descriptions:} \\
            1. Keep it concise. Describe only the main elements, omitting details. \newline
            2. Retain visual elements. Only describe static visuals, avoiding any narrative. \newline
            3. Remove non-visual elements. Typical non-visual elements include dialogue, psychological activities, and plot. \newline
            4. Retain character names. \\
            \textbf{Input Format:} \\
            Input consists of all story content, the current story content, and the corresponding image description for the current story, formatted as: \\
            \{\\
            \hspace{2em}    "all\_stories": ["xxx", "xxx"],\\
            \hspace{2em}    "current\_story": "xxx",\\
            \hspace{2em}    "image\_description": "xxx"\\
            \}\\
            \textbf{Output Format:} \\
            Directly output improvement suggestions, without any additional content. If no improvements are needed, output "Check Passed." \\
            \hline
        \end{tabularx}
    \end{adjustbox}    
    \caption{\label{tab:prompts-image-review} Prompts for image description review based on story descriptions.}
\end{table}

\begin{table*}[t]
    \renewcommand{\arraystretch}{1.5}
    \small
    \begin{adjustbox}{width=\textwidth}
        \begin{tabularx}{\textwidth}{X}
        \hline
            \hline
            \textbf{Invention Story: Understanding Famous Inventors and Their Creations} \\
            \hline
            1. In a small village embraced by the starry sky, there lived a boy named Thomas, who harbored dreams of illuminating the world. \newline 
            2. At the end of the village, an abandoned wooden house became his secret base for achieving his dreams, with moonlight illuminating his path, leading him into the unknown. \newline 
            3. As night fell, the village plunged into tranquility, but inside that wooden house, a glimmer of hope flickered, marking the quiet beginning of Thomas's journey of exploration. \newline
            4. Thomas’s small workshop was filled with various tools and materials, each failure leaving a mark on the walls. \newline 
            5. He discovered that even the smallest experiments had their value; they were like bricks, building the road to success. \newline 
            6. From metals to plant fibers, every possible material was carefully tested by him, and his notebook was filled with dense records of his notes and reflections. \newline 
            7. Late at night, the lab was eerily quiet, with only the faint sounds of machinery. A brave little mouse named Qiqi peeked out from the corner, its eyes sparkling with curiosity. \newline 
            8. During one experiment, Thomas was soaked in sweat, frustrated that a circuit wouldn’t work. Curiously, Qiqi climbed onto the table and accidentally touched a switch; miraculously, the current stabilized. \newline 
            9. Thomas watched in astonishment, suddenly realizing that every partner, no matter how small, could provide unexpected help. He gently patted Qiqi, feeling a surge of gratitude. \newline 
            10. On a pitch-black night, with trembling hands, Thomas carefully placed carbonized bamboo fibers into the experimental light bulb; this was already his 1001st attempt. \newline 
            11. As the current flowed through, a soft, non-blinding light suddenly illuminated the lab, unlike any stable light source he had ever seen. \newline 
            12. The villagers were attracted by this unusual light, stepping out of their homes, amazed at the formerly dilapidated little house that had now become a beacon of hope. \newline 
            13. Thomas took his new invention and visited every corner of the village, personally installing that magical electric light for each household. \newline 
            14. Children sat under the light, their eyes sparkling with the thirst for knowledge, as if they were the stars in the night sky. Now, even the deepest night couldn't hinder their yearning for knowledge. \newline 
            15. The villagers worked under the soft light, laughter and conversations mingling and spreading throughout the community. The light not only illuminated their homes but also warmed their hearts, making the community more harmonious and joyful due to Edison’s invention. \newline 
            16. Night fell, but the world was no longer silent in darkness. From bustling cities to remote villages, each electric light was a gift from Thomas to the world. \newline 
            17. Edison’s name, like his electric light, illuminated the river of human history, becoming a symbol of the pursuit of light and progress. \newline 
            18. “Every great invention begins with a small dream and countless seemingly insignificant efforts.” This sentence inspires future generations to continue exploring the unknown and change the world through innovation. \\
            \hline
        \end{tabularx}
    \end{adjustbox}
    \label{tab:invention-story}
    \caption{ Example Story with the topic ``invention story: understanding famous inventors and their creations''.}
\end{table*}

\end{document}